\documentclass[sigconf,nonacm]{acmart}

\usepackage{graphicx}
\usepackage{subfigure}

\usepackage{amsmath}

\usepackage{savesym}
\savesymbol{Bbbk}
\usepackage{amssymb}
\restoresymbol{AMS}{Bbbk}

\usepackage{mathtools}

\usepackage{arydshln}
\usepackage{booktabs}
\usepackage{multirow}

\usepackage{hyperref}

\usepackage{threeparttable}

\makeatletter
\newcommand*\bigcdot{\mathpalette\bigcdot@{0.8}}
\newcommand*\bigcdot@[2]{\mathbin{\vcenter{\hbox{\scalebox{#2}{$\m@th#1\bullet$}}}}}
\makeatother

\AtBeginDocument{%
  \providecommand\BibTeX{{%
    \normalfont B\kern-0.5em{\scshape i\kern-0.25em b}\kern-0.8em\TeX}}}

\begin{document}

\title{Tel2Veh: Fusion of Telecom Data and Vehicle Flow to Predict Camera-Free Traffic via a Spatio-Temporal Framework}

\author{ChungYi Lin$^{1,2}$, Shen-Lung Tung$^{2}$, Hung-Ting Su$^{1}$, Winston H. Hsu$^{1,3}$}
\affiliation{%
  \institution{ $^{1}$National Taiwan University,    $^{2}$Chunghwa Telecom Laboratories, $^{3}$Mobile Drive Technology
  \city{}
  \country{}
  }
}

\renewcommand{\shortauthors}{ }

\begin{abstract}
 Vehicle flow, a crucial indicator for transportation, is often limited by detector coverage. With the advent of extensive mobile network coverage, we can leverage mobile user activities, or \textit{cellular traffic}, on roadways as a proxy for vehicle flow. However, as counts of cellular traffic may not directly align with vehicle flow due to data from various user types, we present a new \textbf{task}: predicting vehicle flow in \textit{camera-free areas} using cellular traffic. To uncover correlations within multi-source data, we deployed cameras on selected roadways to establish the \textbf{Tel2Veh dataset}, consisting of extensive cellular traffic and sparse vehicle flows. Addressing this challenge, we propose a \textbf{framework} that independently extracts features and integrates them with a graph neural network (GNN)-based fusion to discern disparities, thereby enabling the prediction of unseen vehicle flows using cellular traffic. This work advances the use of telecom data in transportation and pioneers the fusion of telecom and vision-based data, offering solutions for traffic management.
\end{abstract}


\maketitle

\section{Introduction}
\textbf{Vehicle flow} is highly related to traffic conditions, and thus its prediction is useful for reducing congestion and enhancing safety in intelligent transport systems (ITS) \cite{xie2020urban,lv2021deep}. However, its applications are often constrained by the costs and coverage of detectors \cite{zhang2019dynamic}. With the advent of extensive coverage cellular networks \cite{wang2018spatio}, we aim to leverage mobile user network activities (i.e., \textbf{\textit{cellular traffic}} \cite{jiang2022cellular}) collected on roadways to gain insights into traffic conditions. Unlike previous datasets focused on cellular network optimization \cite{barlacchi2015multi,wang2018spatio}, we identify cellular traffic as an emerging resource in ITS.

\begin{figure}[ht]
\centering
\includegraphics[width=0.98\linewidth]{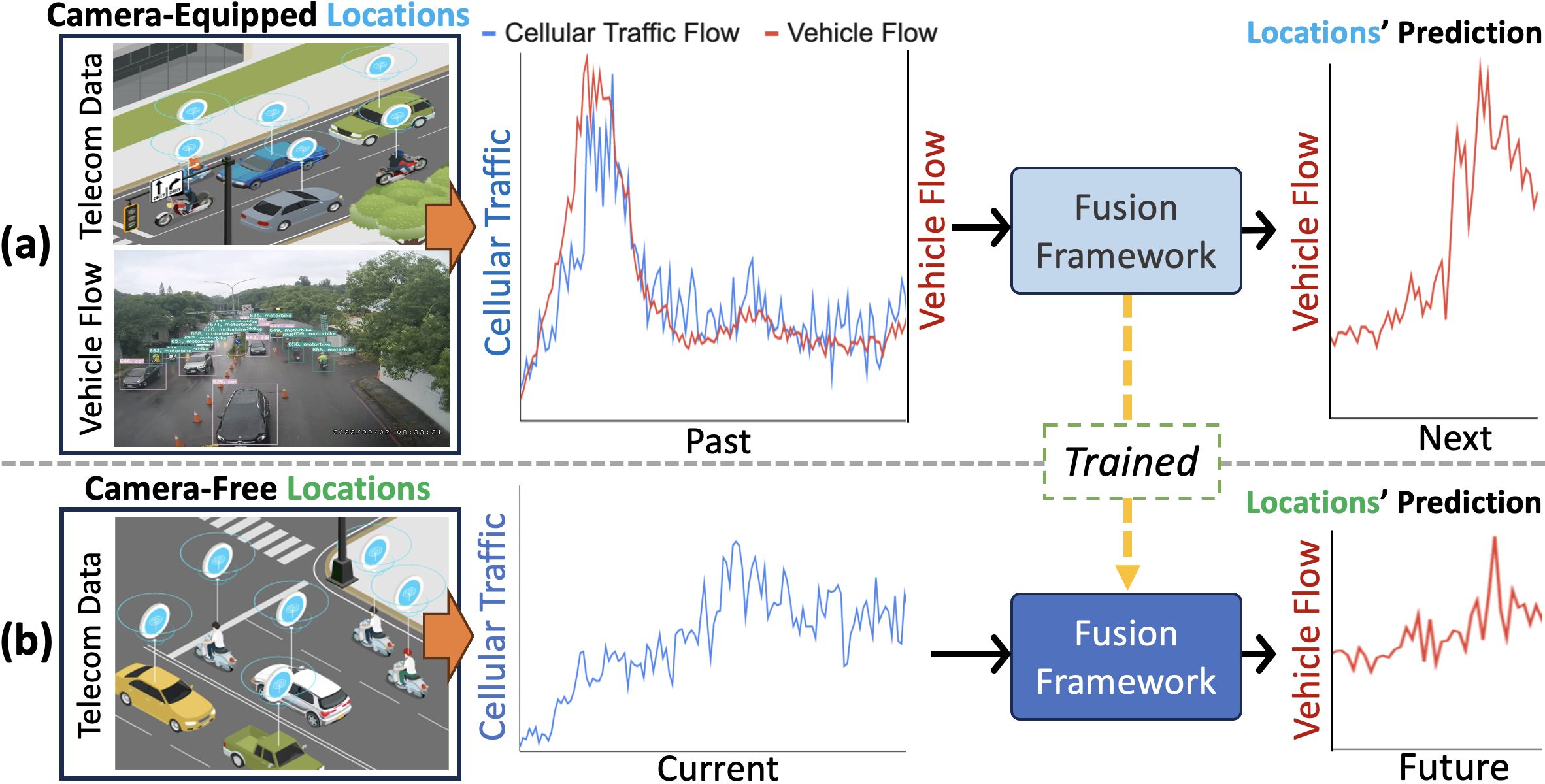}
\caption{Overview of the task and dataset. (a) We propose a task that combines cellular traffic (i.e., mobile user network activities) with sparse vehicle flows to forecast vehicle flow in camera-free areas, supported by the \textbf{Tel2Veh} dataset which contains cellular traffic on roads and camera-detected vehicle flows. (b) Our framework, after training, has proven its capability for accurate prediction in camera-free areas.}
\label{fig:overview}
\end{figure} 

Thus, we further developed \textit{\textbf{Geographical Cellular Traffic (GCT)}}, a type of cellular traffic annotated with its originating location, in collaboration with the leading operator, Chunghwa Telecom. We then aggregated counts of GCT at regular intervals to establish \textit{\textbf{GCT flow}}, revealing the temporal patterns of areas generating GCT. Yet, as the composition of GCT flow reflects a range of network activities from different users like drivers and pedestrians, GCT flow may not accurately represent vehicle flow. Figure \ref{fig:overview}(a) shows similar trends in the temporal patterns of GCT and vehicle flows, while there exists a magnitude disparity between them.

Thus, we present a \textbf{task} of leveraging extensive GCT flows to predict vehicle flows in camera-free areas. We deployed cameras in a few road segments to observe the magnitude disparity between the two flows, resulting in the \textbf{Tel2Veh dataset}, which comprises GCT flows from 49 locations and vehicle flows from 9 of these, covering August to September 2022. To address this, we propose a \textbf{Framework} that operates in two stages: \textit{Stage 1} uses spatial-temporal graph neural networks (STGNNs) for feature extraction from GCT and vehicle flows. \textit{Stage 2} utilizes the GNN-based fusion to integrate these extracted features to predict vehicle flows. Figure \ref{fig:overview}(b) shows that the well-trained framework can use GCT flows to predict the same locations' vehicle flows. This is verified by experiments that accurately predict vehicle flow excluded from training, simulating predictions in camera-free locations.

Overall, our main contributions are as follows:

$\bullet$ \textbf{Novelty}: We pioneer the use of telecom data in ITS and pave the way for fusing telecom and vision-based data sources via our framework for predicting vehicle flows in camera-free areas.

$\bullet$ \textbf{Availability and Utility}: All data and code utilized in this paper are accessible at: \textit{\textbf{\href{https://github.com/cylin-gn/Tel2Veh}{https://github.com/cylin-gn/Tel2Veh}}} and \textit{\textbf{\href{https://doi.org/10.5281/zenodo.10602897}{https://doi.org/10.5281/zenodo.10602897}}}, respectively.

$\bullet$ \textbf{Predicted Impact}: By utilizing cellular traffic data, we offer a cost-effective, scalable alternative for vehicle flow prediction. The Tel2Veh dataset provides a novel resource for further research in ITS, setting a new benchmark in traffic flow prediction. More applications in ITS are detailed in Appendix A.1 on our \textit{{\href{https://cylin-gn.github.io/Tel2Veh-demo}{GitHub}}}.
\section{Tel2Veh Dataset}
The dataset comprises \textbf{GCT} and \textbf{Vehicle Flows}, detailed as follows:

\subsection{Definitions}

\noindent $\bullet$ \textbf{Geographical Cellular Traffic (GCT)}: A cellular traffic record with its originating GPS coordinates, as shown in Table \ref{tab:raw_table}.

\noindent $\bullet$ \textbf{Road Segment}: A 20m x 20m road area, based on Hsinchu City's average road width, serving as the GCT collection boundary.

\noindent $\bullet$ \textbf{GCT Flow}: The cumulative count of GCT within a time interval at the road segment, where each segment exhibits a unique temporal pattern over time, as depicted on the left side of Table \ref{tab:data_structure}.

\noindent $\bullet$ \textbf{Vehicle Flow}: The cumulative count of vehicles detected from the camera within a time interval, as the right side of Table \ref{tab:data_structure}. Details on fine-tuning can be found in Appendix A.2 on our \textit{{\href{https://cylin-gn.github.io/Tel2Veh-demo}{GitHub}}}.

\begin{table}[ht]
    \scriptsize
    \centering
    \begin{minipage}{.42\linewidth}
        \centering
        \captionof{table}{Raw GCT.}
        \label{tab:raw_table}
        \begin{tabular}{|p{0.4cm} p{0.4cm} p{0.5cm} p{0.7cm}|}
            \hline
            \textbf{Time} &
            \textbf{IMEI$^{*}$} &  \textbf{Lat.} &  \textbf{Long.}   \\  \hline
            ... & ...  & ... & ...   \\ 
            07:30:36 & H...aK &  24.78711 & 120.98641  \\ 
            07:31:02 & B...EQ &  24.78702 & 120.98664  \\
           07:31:07 &  M...Gn &  24.78703 & 120.98642  \\
            ... & ...  & ... & ...   \\ \hline
        \end{tabular}
        \scriptsize
  $^{*}$International Mobile Equipment Identity.
    \end{minipage}
    %\hfill
    \begin{minipage}{.53\linewidth}
        \centering
        \captionof{table}{Dataset Structure.}
        \label{tab:data_structure}
        \begin{tabular}{|p{0.3cm} p{0.3cm} p{0.1cm} p{0.3cm}|p{0.6cm} p{0.1cm} p{0.6cm}|}
            \hline
            \textbf{Time} &
            \textbf{1$^{*}$} &  \textbf{...} &  \textbf{49} &
            \textbf{Cam 1} &  \textbf{...} &  \textbf{Cam 9}   \\  \hline
            ... & ...  & ... & ... & ... & ...  & ...     \\ 
            07:30 & 142 & ... & 53 & 656 & ... & 120  \\ 
            07:35 & 158 & ... & 77 & 628 & ... & 135  \\
           07:40 &  177 & ... & 84 & 757 & ... & 185  \\
            ... & ...  & ... & ... & ... & ...  & ...  \\ \hline
        \end{tabular}
        \scriptsize
  $^{*}$ID corresponds to the road segment with GCT flow.
    \end{minipage}

\end{table}

\subsection{Data Collection and Processing}
\noindent \textit{\textbf{GCT Flow:}} 

\begin{figure}[ht]
\centering
\includegraphics[width=0.98\linewidth]{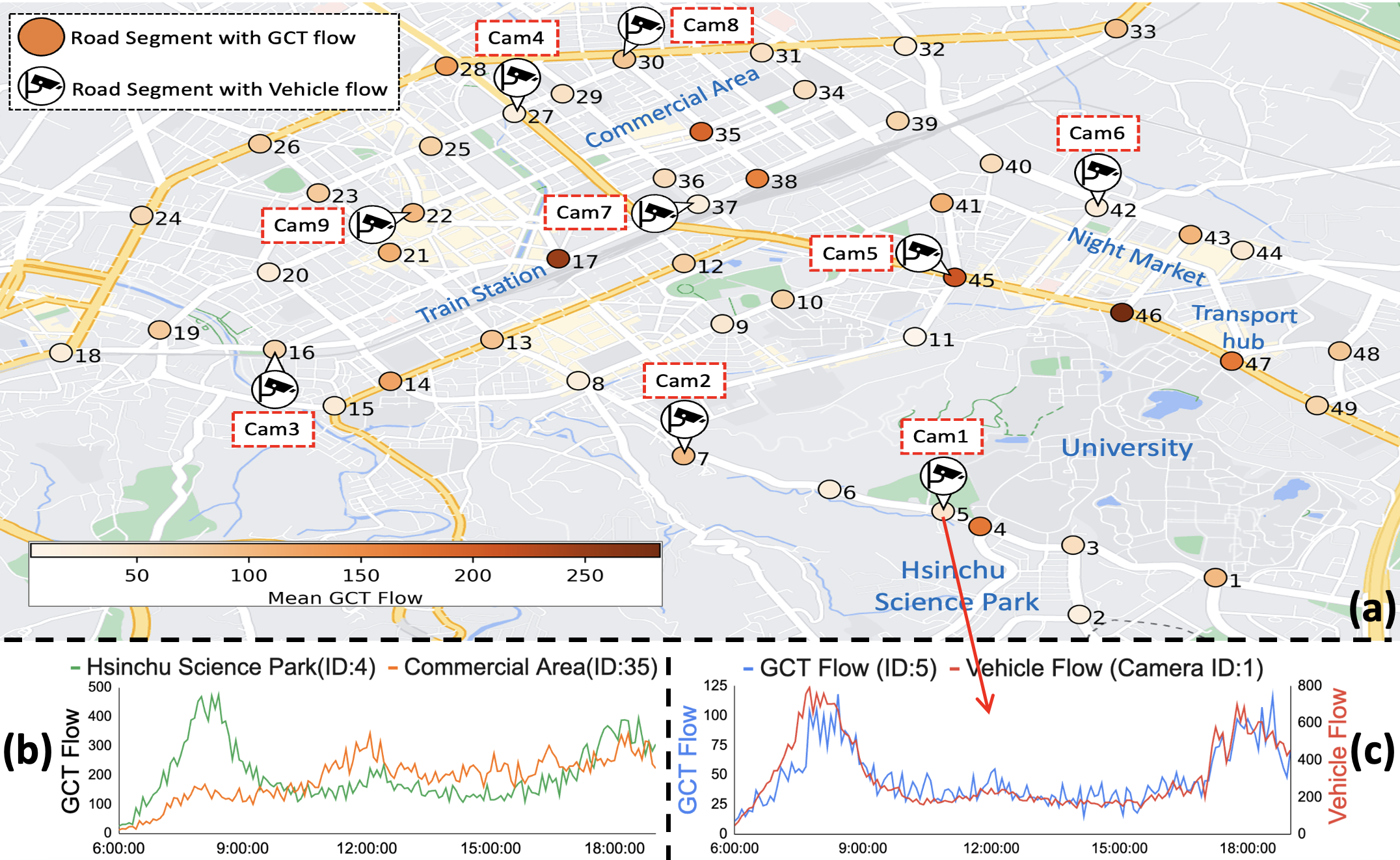}
\caption{Overview of GCT and vehicle flows in Hsinchu City. (a) Spatial distribution of 49 GCT and 9 camera-detected vehicle flows on various road segments. (b) GCT flows show unique temporal dynamics related to functional areas like the Science Park, and Commercial Area. (c) Comparison of GCT flows on Road Segment 5 and vehicle flow from Camera 1 at the same location, highlighting trend similarities.}
\label{fig:road}
\end{figure}

$\bullet$ \textbf{Data Sourcing.} Table \ref{tab:raw_table} was sourced from the telecom company's \textit{Geographical Cellular Traffic Database}. Each row represents a GCT record with essential fields: International Mobile Station Equipment Identity (IMEI), latitude, longitude, and log time. Consulting with \textit{city authorities}, 49 road segments in Hsinchu City were chosen for GCT collection (as Figure \ref{fig:road}), like arterial roads and areas prone to congestion. We accumulated GCTs originating from these segments to obtain GCT flows, as the left side of Table \ref{tab:data_structure}.

$\bullet$ \textbf{Data Privacy.}
To protect user privacy: \textit{(1)} IMEI numbers in  Table \ref{tab:raw_table} were hashed, and personal identifiers like user names and addresses were removed, ensuring anonymity. \textit{(2)} Data collection was confined to roadways, avoiding businesses and residential areas to prevent user tracking. \textit{(3)} The cooperating telecom company adheres to ISO27001 standards, enforcing strict data access approvals.

\noindent \textit{\textbf{Vehicle Flow:}}

$\bullet$ \textbf{Data Sourcing.} We selected 9 road segments (e.g., arterial roads and pedestrian-friendly areas) to deploy cameras, overlapping 9 segments for GCT collection, to detect vehicle flows. The interval for vehicle flow is aligned with that of GCT flow, as the right side of Table \ref{tab:data_structure}. We utilized BOT-Sort \cite{aharon2022bot} and YOLOv7 for tracking and detection, ensuring unique vehicles. Given Taiwan's traffic conditions (e.g., dense motorbike flow), we manually labeled and fine-tuned for accuracy, detailed in Appendix A.2 on our \textit{{\href{https://cylin-gn.github.io/Tel2Veh-demo}{GitHub}}}.

$\bullet$ \textit{\textbf{Flexibility and Scalability.}} Given the extensive mobile network coverage (over 80\% of Taiwan's area), we can \textbf{flexibly} collect GCT from various road segments, aiding in monitoring events or areas lacking detectors, as shown in Figure \ref{fig:road}(a). The flexible collection of GCT flow showcases \textbf{scalability}, expanding our analysis from 10 road segments in our prior work \cite{lin2021multivariate} to the current 49 segments.

$\bullet$ \textit{\textbf{Regional Functionality.}}
Variability in GCT flow over time reflects the unique temporal dynamics of road segments. For instance, Figure \ref{fig:road}(b) shows the Hsinchu Science Park peaking during commuting hours and commercial areas experiencing higher flows at midday, providing valuable insights for traffic pattern analysis.

$\bullet$ \textit{\textbf{Limitations.}} While the trends of GCT and vehicle flows are similar, as shown in Figure \ref{fig:road}(c), there is a magnitude gap between them because our GCT data is sourced from one telecom company.

\noindent {\textbf{Vehicle Flow Strengths and Limitations:}}

$\bullet$ \textit{\textbf{Insights into GCT Flow.}} GCT flow contains diverse user types (drivers, passengers, pedestrians), in contrast to vehicle flow which counts vehicles only. This distinction offers valuable insights into driver behaviors and fundamental vehicular traffic characteristics.

$\bullet$ \textit{\textbf{Assessing Magnitude Disparities.}} Vehicle flow focuses on vehicle count, whereas GCT flow aggregates various mobile user activities, leading to a magnitude disparity between them at identical locations, as depicted in Figure \ref{fig:road}(c). In response, we deploy a few cameras to capture vehicle flows and propose a framework (see Section \ref{framework}) that leverages the magnitude disparity, refining the GCT flow for accurately predicting vehicle flow in camera-free areas.

$\bullet$ \textit{\textbf{Limitations.}} The physical deployment cost constraints limit its flexibility in analyzing especially temporary events.

\begin{table}[ht]
  \scriptsize
  \caption{Descriptive Statistics.}
  \label{tab:statistics}
  \begin{tabular}{lrrrrrr}
    \toprule
    \textbf{Datasets} &  \textbf{$\#$Samples} &  \textbf{$\#$Nodes} &   \textbf{Average} & \textbf{STD}  &  \textbf{Max Avg.} & \textbf{Min Avg.}  \\
    \midrule
    GCT flow & 4240 & 49 & 83.6 & 76.1 & 283.4(ID:46) & 2.7(ID:11) \\
    Vehicle flow & 4240 & 9 & 251.9 & 125.1 & 351.2(Cam5) & 140.3(Cam9) \\
    \bottomrule
    
  \end{tabular}
 
\end{table}
\subsection{Data Analysis}

\noindent $\bullet$ \textit{\textbf{Spatial Coverage.}} For GCT flows, 49 road segments of interest to city authorities were chosen, located near key areas like universities, science parks, and commercial zones. For vehicle flows, cameras were deployed on 9 of these 49 segments to capture vehicle flows. These segments cover congested commute routes (\textit{Cam1, Cam2, Cam3}), land bridge entries (\textit{Cam4, Cam5}), the downtown-highway route (\textit{Cam6, Cam7, Cam8}), and a pedestrian area (\textit{Cam9}).

\noindent  $\bullet$ \textit{\textbf{Temporal Range.}} GCT and vehicle flows were collected from Aug. 28 to Sep. 27, 2022. Counts were accumulated in 5-minute intervals daily from 06:00 to 19:00, targeting high activity hours and yielding a total of 4240 samples, depicted in Table \ref{tab:data_structure}.

\noindent $\bullet$ \textit{\textbf{Descriptive statistics.}} As shown in Table \ref{tab:statistics}, Segment 46, near a university and transfer station, records the highest average GCT flow (283.4), while Segment 11, an alternative route from Science Park, has the lowest (2.7). Cam5, a primary city entry point, records the highest average vehicle flow (351.2). Cam9, situated by a pedestrian zone and off an arterial road, shows the lowest (140.3).

\begin{figure}[ht]
\centering
\includegraphics[width=0.93\linewidth]{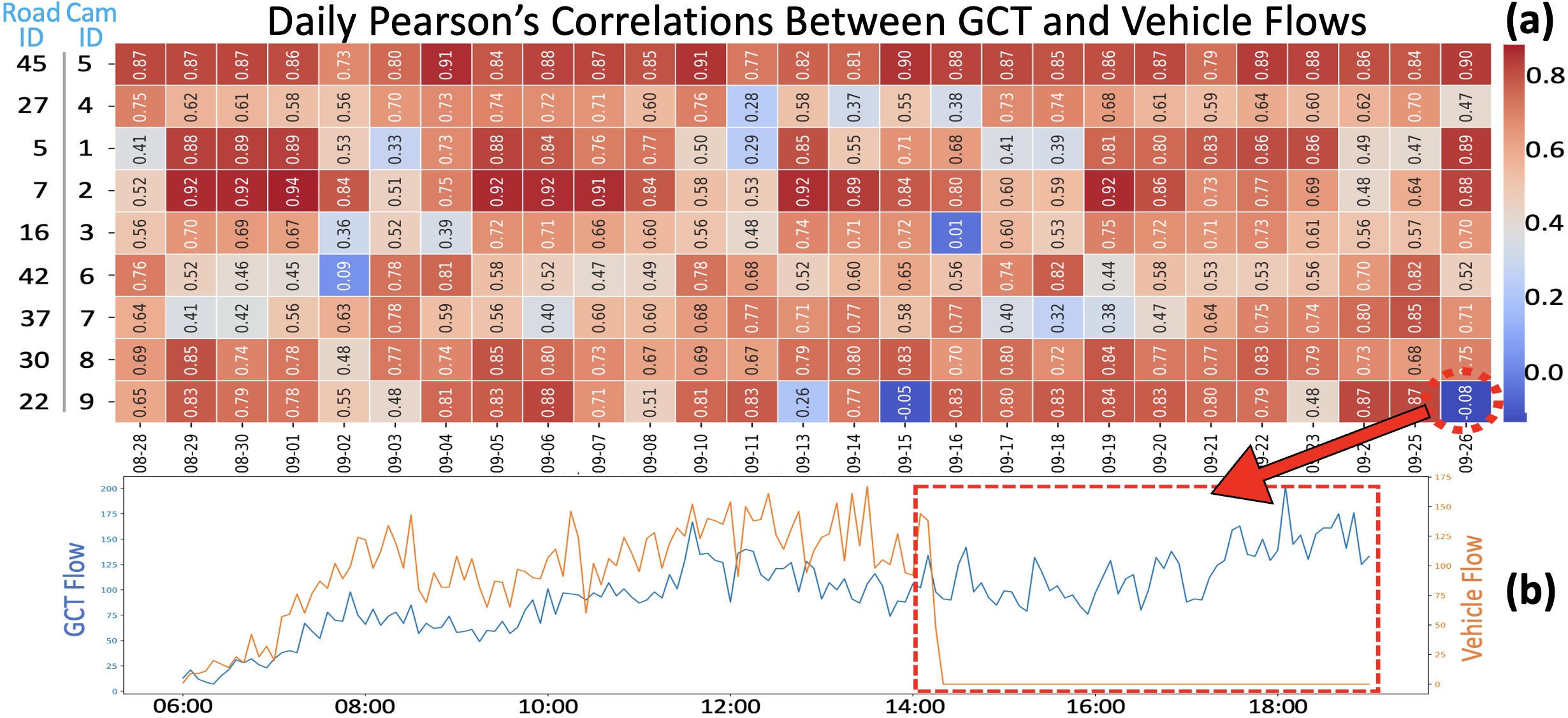}
\caption{Correlation between GCT and vehicle flows. (a) Daily Pearson correlation coefficients at matching locations generally indicate moderate to high correlations, implying pattern alignment, while low correlations suggest data anomalies. (b) An example of low correlation, such as the missing vehicle flow for segment 22 with Cam 9 after 14:00, suggests potential detection errors or device malfunctions.}
\label{fig:pearson_all}
\end{figure}

\noindent $\bullet$ \textit{\textbf{Daily Correlation.}} Figure \ref{fig:pearson_all}(a) displays the daily Pearson correlation coefficient \cite{sedgwick2012pearson} between GCT and vehicle flows at locations (details in Appendix A.3 on our \textit{{\href{https://cylin-gn.github.io/Tel2Veh-demo}{GitHub}}}). The heatmap's grids represent correlations at specific locations and days, with darker oranges indicating stronger correlations. Most locations exhibit moderate to high correlations, implying similar traffic patterns, while low correlations could indicate data anomalies, as Figure \ref{fig:pearson_all}(b).

\section{Spatio-Temporal Fusion Framework}
\label{framework}

\noindent \textbf{Task Definition.} Given \textbf{$N$} GCT flows and \textbf{$M$} vehicle flows from past intervals ($T_{in}$), predict \textbf{$N$} future vehicle flows for upcoming intervals ($T_{out}$), including those in camera-free areas, where $N > M$.

\begin{figure}[ht]
\centering
\includegraphics[width=0.93\linewidth]{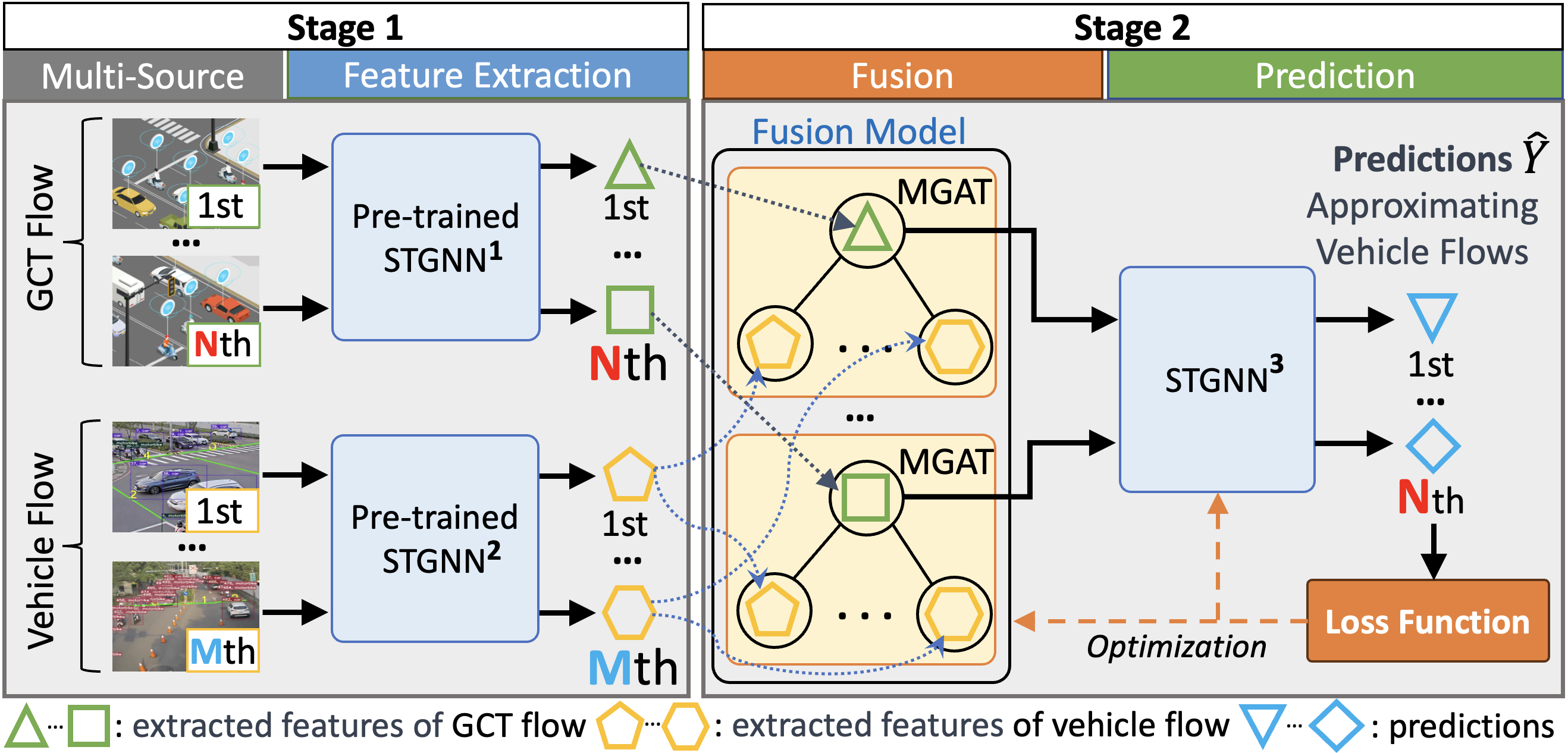}
\caption{Overview of the two-stage fusion framework. Stage 1 uses two pre-trained STGNNs for feature extraction from GCT and vehicle flows. Stage 2 integrates the extracted features via the Fusion Model, feeding them into the third STGNN for prediction. Only Stage 2 is trained, optimizing the Loss Function to align predictions with vehicle flows.}
\label{fig:fusion_model}
\end{figure}

\noindent \textbf{Overview.} Our framework involves a two-stage process, as Figure \ref{fig:fusion_model}. \textit{Stage 1} employs two pre-trained STGNN models to extract features from GCT and vehicle flows separately. \textit{Stage 2} utilizes a GNN-based fusion model to integrate these extracted features, subsequently feeding the results to the third STGNN for vehicle flow prediction. \textit{Stage 2} is trained with the proposed \textit{Loss Function}, dynamically balancing losses from predictions against vehicle flows in camera-equipped areas and against GCT flows in camera-free areas.

\noindent \textbf{Stage 1: Feature Extraction.} We pre-trained\footnote{For details on training STGNN, see:\href{https://github.com/nnzhan/Graph-WaveNet}{https://github.com/nnzhan/Graph-WaveNet}} STGNNs on forecasting GCT and vehicle flows separately, extracting their spatio-temporal features due to their computational efficiency \cite{wu2019graph}. Current STGNNs \cite{wu2019graph,ye2022learning,lin2024teltrans} use $K$ kernels in a 1D convolution network to convert spatio-temporal input (e.g., $N$ locations with $D$ observations, size [$N \times D$]) into \textit{multi-channel feature maps} (size [$K \times N \times D$]), where each map could encompass distinct temporal patterns. We then utilize these extracted multi-channel feature maps for Stage 2.

\noindent \textbf{Stage 2: Fusion and Prediction:} 

\textbf{\textit{Preliminary of Fusion.}} 
For data fusion, the Graph Attention Network (GAT) is ideal due to its ability to dynamically adjust weights among features \cite{zhao2020multivariate}. However, when fusing multi-channel feature maps, existing GAT implementations that assign uniform weight across feature maps may overlook critical patterns in specific feature maps, potentially leading to suboptimal results \cite{brody2022how}. Thus, we propose a method for fusing such format features as follows.

\textbf{\textit{Fusion.}} To handle multi-channel feature maps, we adopt Multi-channel Graph Attention (\textbf{MGAT}) from our prior work \cite{lin2023pay} (summarized in Appendix A.4 on \textit{{\href{https://cylin-gn.github.io/Tel2Veh-demo}{GitHub}}}) as the basis of our fusion model. Each \textit{GCT flow feature} is fused with all $M$ \textit{vehicle flow features}:
\begin{equation}
\hat{H}^{gct}_{n}=MGAT(\{H^{gct}_{n}, H^{veh}_{1}, \dots , H^{veh}_{M}\},\mathcal{G}),
\label{eq:fusion_model_1}
\end{equation}
where $H^{gct}_{n}$ and $(H^{veh}_{1}, \dots, H^{veh}_{M})$ are the features of the $n$-th GCT flow and the 1st to $M$-th vehicle flows, respectively. $\mathcal{G}$ is the adjacency graph among the corresponding locations of features.

\textbf{\textit{Prediction.}}
After processing all $N$ GCT flow features with Equation \ref{eq:fusion_model_1}, the enhanced features (i.e., $\hat{H}^{gct}_{1},\dots,\hat{H}^{gct}_{N}$) are input into the third STGNN (STGNN$^{3}$) for spatio-temporal modeling, generating $N$ predictions closely approximating vehicle flows.

\textbf{\textit{Loss Function.}} For Stage 2 training, we designed a loss function with two components: one evaluates predictions against actual vehicle flows, and the other assesses predictions against GCT flows. Both components utilize Mean Absolute Error (MAE) for evaluation:

$\bullet$ $\mathcal{L}_{w}$: Loss between observed vehicle flows ($Y_{veh}$) at locations \textit{\textbf{with}} cameras and their predictions ($\hat{Y}_w$):
\begin{math}
\mathcal{L}_{w} = MAE(Y_{veh}, \hat{Y}_{w}),
\end{math}
specifying predictions at camera-equipped locations.

$\bullet$ $\mathcal{L}_{w/o}$: Loss for GCT flows ($Y_{gct}$) at locations \textit{\textbf{without}} cameras against to their predictions ($\hat{Y}_{w/o}$):
\begin{math}
\mathcal{L}_{w/o} = MAE(Y_{gct}, \hat{Y}_{w/o}) ,
\end{math}
targeting predictions at camera-free locations.

Notably, \textbf{the overall predictions, $\hat{Y}$}, are a concatenation of $\hat{Y}_{w}$ and $\hat{Y}_{w/o}$. Thus, the loss function is defined as:
\begin{equation}
\mathcal{L} = \mathcal{L}_{w} + \lambda \mathcal{L}_{w/o} \, ,
\label{eq:dynamic_loss}
\end{equation}
where $\lambda$ is a learnable parameter updated during training to balance $\mathcal{L}_{w}$ and $\mathcal{L}_{w/o}$. We minimize the loss function $\mathcal{L}$ by updating parameters in Stage 2, including those in the fusion model and the third STGNN, to optimize the framework's predictive accuracy.
\section{Experiments}
\subsection{Experimental Settings and Baselines}

\noindent  \textbf{Metrics.} Following \cite{wu2019graph}, we employ Mean Absolute Error (MAE), Root Mean Squared Error (RMSE), and Mean Absolute Percentage Error (MAPE) to evaluate our predictions against real data.

\noindent  \textbf{Baselines.} 
We selected recent STGNN models to integrate within our framework: \textbf{\textit{GWNET}} \cite{wu2019graph}: A GNN-based Wavenet with a spatial diffusion mechanism. \textbf{\textit{ESG}} \cite{ye2022learning}: A GNN-based model employing evolutionary and multi-scale graph structures. \textbf{\textit{MFGM}} \cite{lin2024teltrans}: A GNN-based model capturing multivariate, temporal, and spatial dynamics.

\noindent  \textbf{Settings.} Baselines are evaluated \textbf{without(w/o)} and \textbf{with(w)} integration into our framework, both trained with the proposed Loss Function $\mathcal{L}$, with the $\lambda$ initialized from 0.0001 to 0.00001. \textit{\textbf{Data Setups}} and \textit{\textbf{Model Settings}} are in Appendix A.5 on our \textit{{\href{https://cylin-gn.github.io/Tel2Veh-demo}{GitHub}}}.

\begin{table}[ht]
\centering
\caption{Models of With/Without Framework Integration.}
\label{pred_table}
\begin{threeparttable}
\footnotesize
\setlength\tabcolsep{1.5pt}
\begin{tabular*}{\columnwidth}{@{\extracolsep{\fill}} l ccc ccc ccc}
    \toprule
      & \multicolumn{3}{c}{15 mins.} & 
        \multicolumn{3}{c}{30 mins.} & 
        \multicolumn{3}{c}{60 mins.} \\ 
    \cmidrule{2-10}
    \textit{Baselines}  & MAE & RMSE & MAPE
      & MAE & RMSE & MAPE
      & MAE & RMSE & MAPE 
      
      \\
    \midrule
     
    GWNET(w/o$^{1}$)
    
    & 116.7 & 156.5 & 44.6\% 
    & 122.6 & 158.1 & 46.9\% 
    
    & 130.8 & 168.1 & 50.2\% \\
     GWNET(w)
    & \textbf{103.6} & \textbf{125.7} & \textbf{39.4\%} 
    & \textbf{105.6} & \textbf{126.6} & \textbf{40.5\%} 
    & \textbf{106.5} & \textbf{127.3} & \textbf{42.7\%} \\
    \hdashline
    \textit{IR$^{2}$} 
    & 11.2\% & 19.7\% &	11.7\%	
    & 13.9\% &	19.9\% &	13.6\%	
    & 18.6\% &	24.3\% &	14.9\% \\
    
    \midrule

    ESG(w/o)
    & 108.6 & 141.8 & 45.5\% 
    & 112.5 & 146.1 & 46.1\% 
    & 126.7 & 163.8 & 47.6\% \\
    ESG(w)
    & \textbf{97.32} & \textbf{120.4} & \textbf{36.5\%} 
    & \textbf{99.38} & \textbf{123.1} & \textbf{36.9\%} 
    & \textbf{100.9} & \textbf{124.4} & \textbf{37.6\%} \\
    \hdashline
    \textit{IR} 
    & 10.4\% &	15.1\% &	19.8\% 	
    & 11.7\% &	15.7\% &	20.9\% 	
    & 20.4\% &	24.1\% &	21.1\%  \\
    
     \midrule
    MFGM(w/o)
    
    & 89.67 & 123.3 & 37.2\% 
    & 91.35 & 124.6 & 38.7\% 
    & 101.01 & 137.5 & 43.7\% \\
    MFGM(w)  
    & \textbf{73.37} & \textbf{98.91} & \textbf{28.9\%} 
    & \textbf{74.89} & \textbf{99.73} & \textbf{29.9\%} 
    & \textbf{77.04} & \textbf{102.9} & \textbf{31.3\%} \\
    \hdashline
    \textit{IR} 
    & 18.2\% &	19.8\%	& 22.3\%	
    & 18.0\% &	19.9\%	& 22.7\%	
    & 23.7\% &	25.2\%	& 28.4\%  \\
    \midrule
    Average IR 
   
     & 13.3\%	 & 18.2\%	& 17.9\%	
     & 14.5\%	 & 18.5\%	& 19.1\%	
     & 20.9\%	 & 24.5\%	& 21.4\% \\

    \bottomrule
\end{tabular*}
\begin{tablenotes}\scriptsize
\item $^{1}$The model is used for prediction without (w/o) integration into our framework.
\item $^{2}$IR (Improvement Ratio) is calculated by ((score(w) - score(w/o)) / score(w/o)) * 100\%.
\end{tablenotes}
\end{threeparttable}
\end{table}

\subsection{Experimental Results}
\noindent {\textbf{Camera-Free Scenario.}} For each configuration in Table \ref{pred_table}, we \textit{\textbf{sequentially excluded one vehicle flow}} per training and used this excluded flow as the ground truth to assess the model's predictions. The results from all exclusion trainings were then averaged as the model's ability in unseen scenarios. For reliability, we run \textit{\textbf{10 times}} for each configuration, with the mean results in Table \ref{pred_table}.

\noindent {\textbf{Prediction Improvement.}} Table \ref{pred_table} offers short-to-long term forecasts, showing that integrating models into our framework significantly improves accuracy, as evidenced by the Improvement Ratio (\textit{\textbf{IR}}). Among them, MFGM achieves superiority and the highest IR, with the MAPE's IR from 22.3\% in short-term to 28.4\% in long-term.

As prediction intervals lengthen, accuracy usually drops due to complex long-term dependencies. Yet, our framework consistently enhances accuracy across all models and yields progressively larger improvements over longer forecasts, as the rising \textit{\textbf{average IR}} shows. This confirms our framework's efficacy in enhancing STGNNs for complex, long-term forecasts, highlighting its practical value.

\begin{figure}[ht]
\centering
\includegraphics[width=0.98\linewidth]{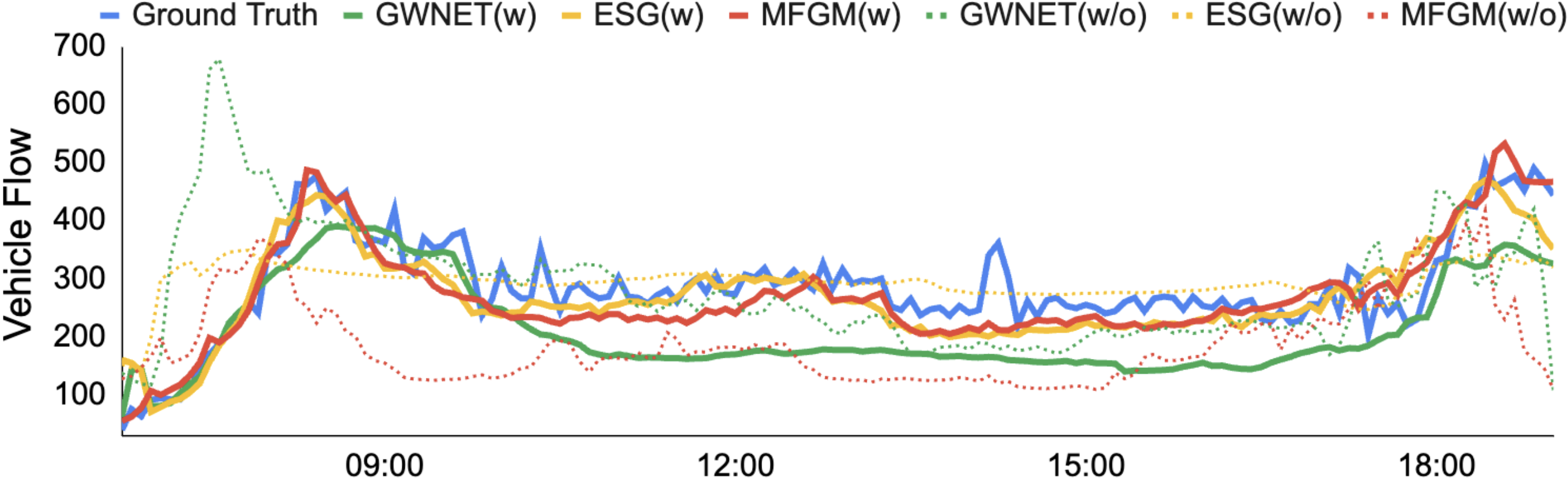}
\caption{Daily 15-minute forecasts on Road ID 37 for 9/26. Ground truth (i.e., actual vehicle flow from Cam7) was excluded from training to serve as validation. The closer alignment of solid lines with the ground truth confirms the accuracy enhancement by our framework.}
\label{fig:line_graph}
\end{figure}

Figure \ref{fig:line_graph} visualizes daily 15-minute predictions, comparing models with (solid lines) and without (dotted lines) framework integration. Solid lines align more closely with the ground truth (i.e., actual vehicle flow) and peak timings, highlighting our framework's ability to address magnitude disparity and capture temporal trends, supporting the improving forecasting results in Table \ref{pred_table}.

\section{Conclusion}
We present a novel resource, \textit{\textbf{Tel2Veh}}, to facilitate the \textit{\textbf{task}} of using telecom data to predict vehicle flows in camera-free areas. Thus, we propose a \textit{\textbf{fusion framework}} that integrates multi-source data for accurate predictions in unseen scenarios. Our work advances multi-source fusion and the application of telecom data in transportation. 

\begin{acks}
This work was supported in part by National Science and Technology Council, Taiwan, under Grant NSTC 111-2634- F-002-022 and by Qualcomm through a Taiwan University Research Collaboration Project.
\end{acks}

\bibliographystyle{ACM-Reference-Format}
\bibliography{sample-base}

\clearpage

\appendix

\end{document}